\documentclass[sigconf]{acmart}

\AtBeginDocument{%
  \providecommand\BibTeX{{%
    \normalfont B\kern-0.5em{\scshape i\kern-0.25em b}\kern-0.8em\TeX}}}

\copyrightyear{2023} 
\acmYear{2023} 
\setcopyright{acmlicensed}\acmConference[CIKM '23]{Proceedings of the 32nd ACM International Conference on Information and Knowledge Management}{October 21--25, 2023}{Birmingham, United Kingdom}
\acmBooktitle{Proceedings of the 32nd ACM International Conference on Information and Knowledge Management (CIKM '23), October 21--25, 2023, Birmingham, United Kingdom}
\acmPrice{15.00}
\acmDOI{10.1145/3583780.3615055}
\acmISBN{979-8-4007-0124-5/23/10}

\usepackage{subfigure}
\usepackage{bm}
\usepackage{color, soul}
\definecolor{codegray}{rgb}{0.5,0.5,0.5}

\usepackage{multirow}
\usepackage{bm}
\usepackage{enumitem}
\usepackage{amsmath}
\usepackage{makecell}

\usepackage[linesnumbered,vlined,ruled,noend]{algorithm2e}

\begin{document}

\title{Semantic-aware Node Synthesis for Imbalanced Heterogeneous Information Networks}

\author{Xinyi Gao}
\affiliation{
\institution{The University of Queensland}
\city{Brisbane}
\country{Australia}
}
\email{xinyi.gao@uq.edu.au}

\author{Wentao Zhang}
\affiliation{
\institution{Peking University}  
\city{Beijing}
\country{China}
}
\email{wentao.zhang@pku.edu.cn}

\author{Tong Chen}
\affiliation{
\institution{The University of Queensland}
\city{Brisbane}
\country{Australia}
}
\email{tong.chen@uq.edu.au}

\author{Junliang Yu}
\affiliation{
\institution{The University of Queensland}
\city{Brisbane}
\country{Australia}
}
\email{jl.yu@uq.edu.au}


\author{Hung Quoc Viet Nguyen}
\affiliation{
\institution{Griffith University}
\city{Gold Coast}
\country{Australia}
}
\email{henry.nguyen@griffith.edu.au}


\author{Hongzhi Yin}
\authornote{This author is the corresponding author.}
\affiliation{
\institution{The University of Queensland}  
\city{Brisbane}
\country{Australia}
}
\email{h.yin1@uq.edu.au}

\begin{abstract}
Heterogeneous graph neural networks (HGNNs) have exhibited exceptional efficacy in modeling the complex heterogeneity in heterogeneous information networks (HINs). The critical advantage of HGNNs is their ability to handle diverse node and edge types in HINs by extracting and utilizing the abundant semantic information for effective representation learning. As a widespread phenomenon in many real-world scenarios, the class-imbalance distribution in HINs creates a performance bottleneck for existing HGNNs. Apart from the node imbalance in quantity, the more crucial and distinctive challenge in HINs is \textit{semantic imbalance}. Minority classes in HINs often lack diverse and sufficient neighbor nodes, resulting in biased and incomplete semantic information. This semantic imbalance further compounds the difficulty of accurately classifying minority nodes, leading to the performance degradation of HGNNs. However, existing remedies are either tailored for non-graph data or designed specifically for homogeneous graphs, thus overlooking the inherent semantic imbalance in HINs. To tackle the imbalance of minority classes and supplement their inadequate semantics, we present the first method for the semantic imbalance problem in imbalanced HINs named Semantic-aware Node Synthesis (SNS). By assessing the influence on minority classes, SNS adaptively selects the heterogeneous neighbor nodes and augments the network with synthetic nodes while preserving the minority semantics. In addition, we introduce two dedicated regularization approaches for HGNNs that explore the inter-type and intra-type information and constrain the representation of synthetic nodes from both semantic and class perspectives to effectively suppress the potential noises from synthetic nodes, facilitating more expressive embeddings for classification. The comprehensive experimental study demonstrates that SNS consistently outperforms existing methods in different benchmark datasets. 
\end{abstract}

\ccsdesc[500]{Computing methodologies~Artificial intelligence}
\ccsdesc[500]{Mathematics of computing~Graph theory}

\keywords{Heterogeneous information networks, semantic imbalance, imbalanced node classification, data augmentation, semantic-aware neighbor selection.}

\maketitle
\section{Introduction}

\begin{figure}[t]
\setlength{\abovecaptionskip}{0.2cm}
\centering
\includegraphics[width=\linewidth]{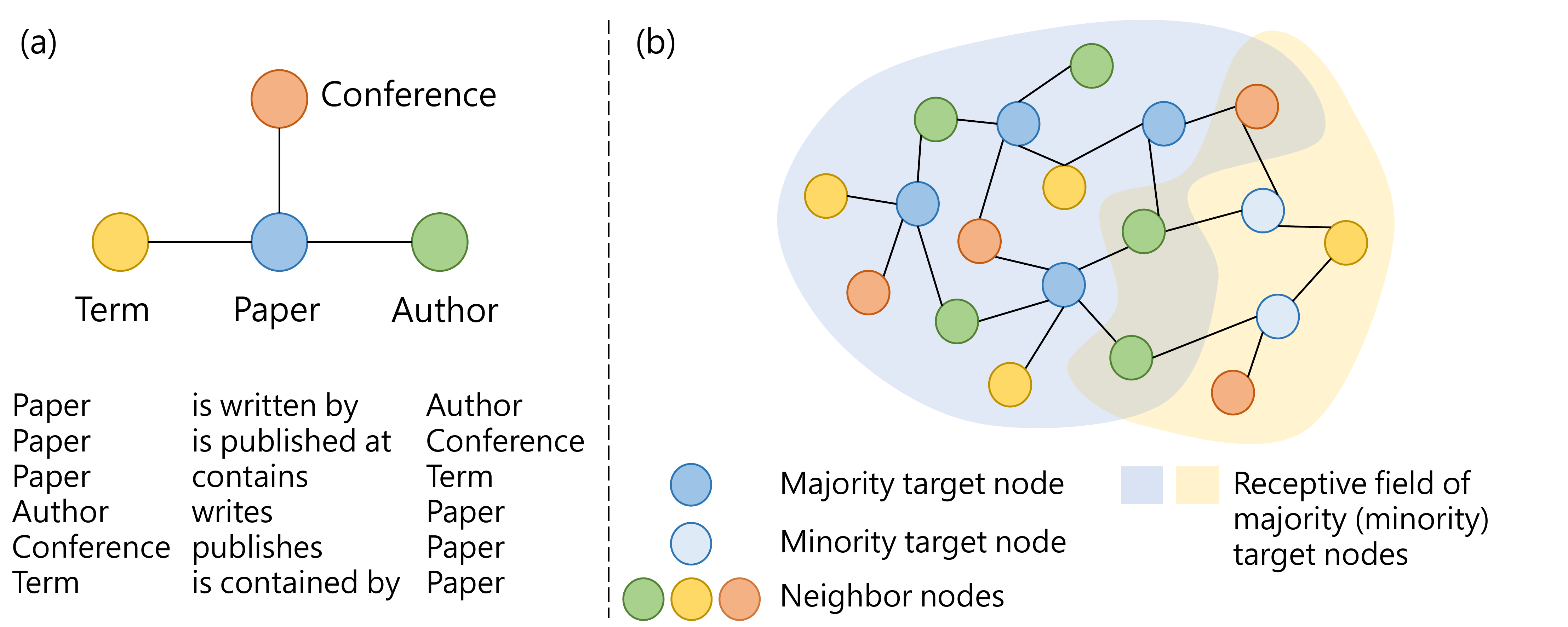}
\caption{(a) The network schema and relations of heterogeneous information network DBLP. (b) The semantic imbalance problem between the majority class and minority class for target nodes. The receptive field indicates the range of first-order neighbors for majority (minority) target nodes.
}
\label{fig_pre}
\end{figure}

In the context of big data, heterogeneous information networks (HINs) also known as heterogeneous graphs play a pivotal role in composing different types of nodes (a.k.a., objects) and relations, and they have become ubiquitous in real-world scenarios, ranging from bibliography networks, social networks to recommender systems \cite{wang2022survey,wu2022graph,2017SPTF,yu2021socially,2017Argument, chen2020try, 10144391}. As shown in Fig. \ref{fig_pre} (a), HINs consist of multiple types of nodes and relations that jointly provide rich semantics.
Recently, heterogeneous graph neural networks (HGNNs) \cite{9300240, zhu2019relation, hong2020attention, hu2020heterogeneous, lv2021we, yang2022simple, yang2021interpretable,sun2021heterogeneous} have been widely used to model the complex heterogeneity and rich semantic information in HINs, and they have made great success in a variety of downstream tasks, such as node classification and link prediction. To achieve competitive predictive performance, HGNNs generally assume abundant and balanced task-specific labeled data. 
Unfortunately, this assumption is hard to sustain due to the time-consuming and resource-intensive data annotation process. In many real-world scenarios, node classes and labels exhibit an imbalanced distribution in HINs, i.e., some classes have significantly fewer samples than other classes.

Take graph-based anomaly detection as an example, nodes that are labeled as abnormal are commonly the minority compared with benign ones. Besides, in the context of paper topic classification in bibliography networks, there may be limited resources available for obsolete topics compared to on-trend topics.

\begin{figure}[t]
\setlength{\abovecaptionskip}{0.2cm}
\centering
\begin{minipage}[t]{0.47\linewidth}
\centering
\includegraphics[width=\linewidth]{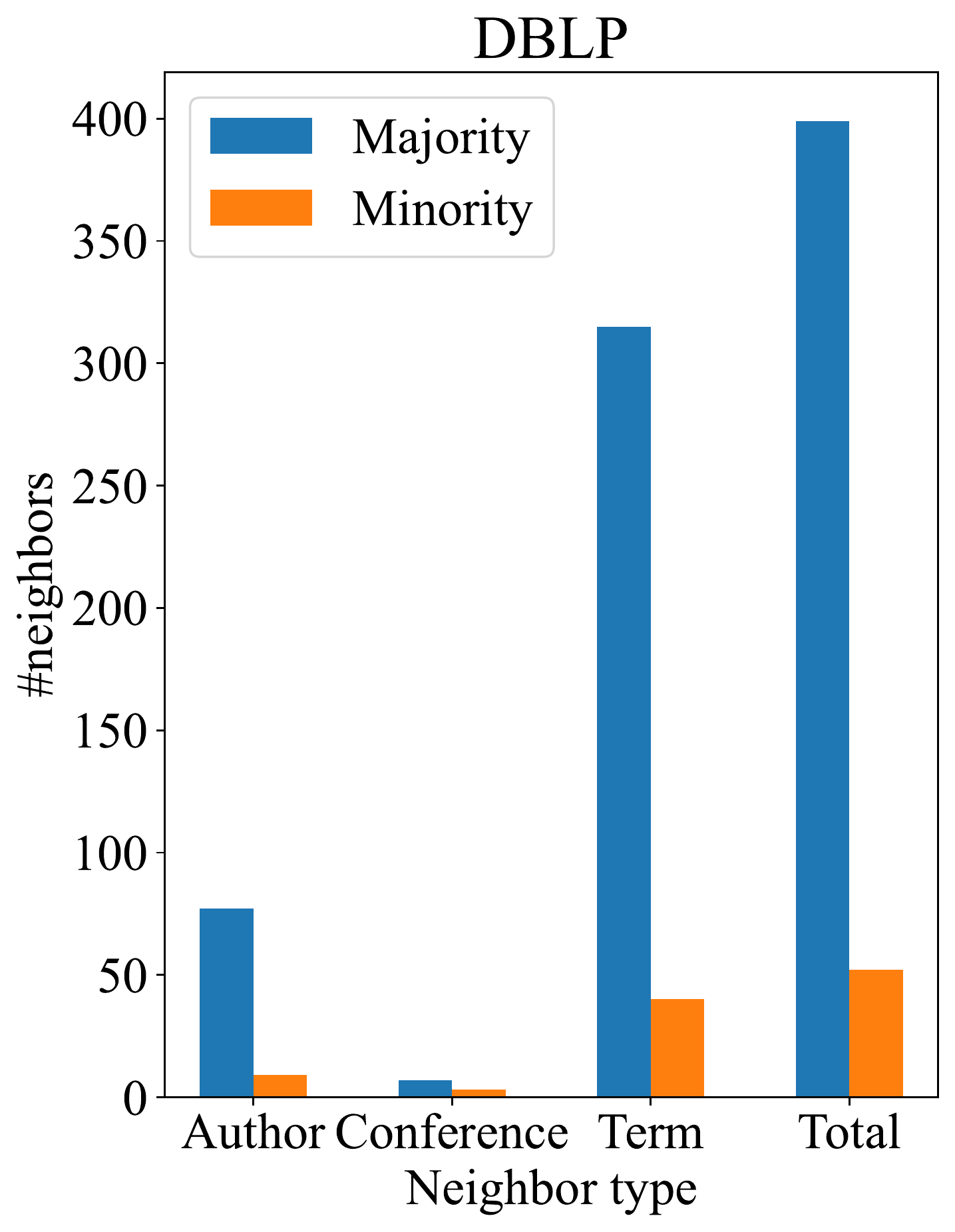}
\end{minipage}
\begin{minipage}[t]{0.47\linewidth}
\centering
\includegraphics[width=\linewidth]{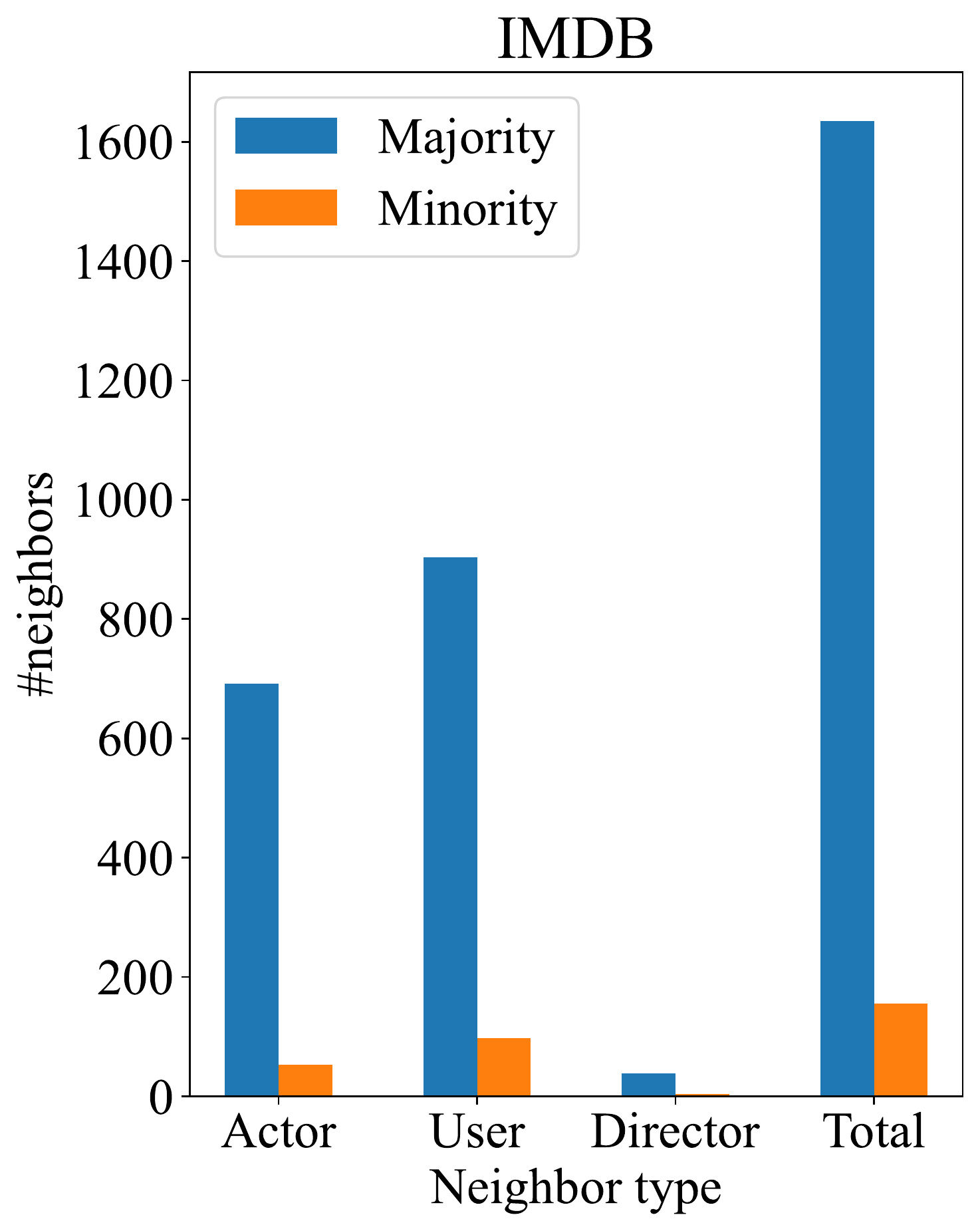}
\end{minipage}
\caption{Statistics of first-order neighbor nodes for majority and minority in DBLP and IMDB dataset. The target node types are paper and movie, respectively. The label rate is 3\% and the imbalance ratio is 0.1.
}
\label{fig_obs}
\end{figure}

Compared with homogeneous graphs \cite{zhang2021rim,zhang2022graph,DBLP:conf/sigmod/ZhangSLCY021,hung2017computing,nguyen2017argument}, on top of the quantity imbalance of nodes, HINs exhibit a unique type of imbalance: \textit{semantic imbalance} which seriously impairs the performance of HGNNs.
Specifically, given the information propagation nature of HGNNs, the class for each node in HINs is no longer simply associated with its own attributes but is also strongly impacted by the semantics carried by various neighbor nodes. 
In HINs, nodes within a minority class usually suffer from a below-average quantity and diversity of their neighbor nodes, where the lack of semantic information heavily impairs the quality of their learned representations.
As shown in Fig. \ref{fig_pre} (b), in the task of paper topic classification within a bibliography network, papers working on minority topics face inevitable information scarcity of authors, conferences, and terms that is highly related to the minority topics. 
Furthermore, the statistical results for DBLP and IMDB datasets \cite{yang2022self} in Fig. \ref{fig_obs} show that the volume of neighbor nodes is highly imbalanced between minority and majority nodes, and the observation is consistent w.r.t. different types of neighbor nodes. Due to the message passing across heterogeneous neighbors in HGNNs, this imbalance in neighbor semantics will exacerbate the under-representation issue of minority nodes.

Intuitively, the lack of an adequate quantity of minority neighbors is the underlying cause of the semantic imbalance problem in HINs. However, complementing the minority neighbors poses a complex and challenging task. One promising solution to information complementation is data augmentation  \cite{chawla2002smote,zhao2021graphsmote,qu2021imgagn}, which generates pseudo nodes and edges for minorities to re-balance the data and provide additional training information. However, simply increasing the number of nodes and neighbors may not yield satisfactory results in HINs since it fails to comprehensively consider the semantic roles played by the generated pseudo nodes and edges. As a result, the neighborhood information that comes with the generated pseudo nodes is still unable to fill the semantic gap for minority classes. 
To assess the effect of neighbor quality, we simulate node augmentation for imbalanced node classification in HINs while controlling for attribute influence.
The imbalanced binary classification task is constructed on DBLP and IMDB datasets, where the dataset settings are the same as Fig. \ref{fig_obs}. The label rate is 3\% and imbalance ratio=0.1 (see Section \ref{preliminary} for definition)\footnote{As a proof-of-concept, we hold out real minority nodes and directly use them as the pseudo nodes during the augmentation process. This is slightly different from our main experiments' setting as we would like to make this discussion independent from the quality of generated pseudo minority nodes.}. Among the three augmentation strategies, $AUG_{ran}$ random links the pseudo nodes with other types of neighbor nodes, $AUG_{min}$ only links the pseudo nodes with the neighbor nodes of minority nodes, and $AUG_{bal}$ mimics the optimal augmented HIN by adding back the original edges of pseudo nodes. According to the classification results in Table \ref{tab_ob1}, all three data augmentation methods improve the imbalanced classification performance by addressing quantity imbalance. However, both $AUG_{ran}$ and $AUG_{min}$ led to performance degradation compared with $AUG_{bal}$. As a note on graph semantics, $AUG_{ran}$ and $AUG_{min}$ respectively alter the original distribution and reduce the diversity of the neighbors w.r.t. the minority nodes, exacerbating the under-representation issue of minority nodes and leading to sub-optimal classification results. In light of these findings, the key question that arises for imbalanced node classification in HINs is: \textit{how to simultaneously complement minority classes in quantity and supplement realistic semantic information in HINs?}

\begin{table}[t]
\setlength{\abovecaptionskip}{0.2cm}
\caption{{The binary classification results of DBLP and IMDB for different neighbor selection methods. Base model is ie-HGCN \cite{yang2021interpretable}.}}
\label{tab_ob1}
\resizebox{\linewidth}{!}{
\begin{tabular}{l|lll|lll}
\hline
          & \multicolumn{3}{c|}{DBLP}            & \multicolumn{3}{c}{IMDB}             \\ \hline
Method    & F1 score   & ACC        & BACC       & F1 score   & ACC        & BACC       \\ \hline
Imbalance & 91.62$\pm$1.78 & 91.62$\pm$1.78 & 91.82$\pm$1.7  & 49.74$\pm$0.16 & 50.64$\pm$0.32 & 50.04$\pm$0.42 \\
$AUG_{ran}$    & 93.68$\pm$1.08 & 93.69$\pm$1.08 & 93.81$\pm$1.01 & 51.55$\pm$0.96 & 52.57$\pm$3.19 & 52.98$\pm$0.75 \\
$AUG_{min}$  & 93.75$\pm$1.15 & 93.75$\pm$1.15 & 93.91$\pm$1.09 & 51.31$\pm$1.1  & 52.05$\pm$1.71 & 53.33$\pm$1.64 \\
$AUG_{bal}$   & 94.54$\pm$1.23 & 94.55$\pm$1.23 & 94.61$\pm$1.22 & 52.61$\pm$1.82 & 54.36$\pm$1.17 & 54.06$\pm$1.78 \\ \hline
\end{tabular}}
\end{table}

While classification under data imbalance has been well-studied previously \cite{sun2009classification, haixiang2017learning, johnson2019survey}, existing methods, even those designed for homogeneous graphs, fail to handle the heterogeneous semantics and structural properties that are intrinsic and essential in HINs, making this problem still under-explored.
For example, motivated by the classical SMOTE algorithm \cite{chawla2002smote}, GraphSMOTE \cite{zhao2021graphsmote} leverages the graph neural networks to learn both topological structures and node embeddings and re-balances the graph by generating synthetic nodes. 
ImGAGN \cite{qu2021imgagn} introduces GAN into the link generation task and synthesizes the minority by interpolating existing minority nodes.
GraphENS \cite{park2022graphens} creates the ego network for each synthetic node by comparing the embeddings for two existing ego networks.
However, these methods are developed to address the issue of node class imbalance in homogeneous graphs, where all nodes belong to the same type. In contrast, HINs consist of diverse node and edge types, with the homogeneous graphs studied in these works considered as simplistic subgraphs or special cases within the broader HIN framework. Consequently, these methods are inadequate in handling the diverse neighbor types and the associated semantic information, particularly the problem of semantic imbalance, are thus inapplicable or sub-optimal for HINs.

In this paper, we aim to tackle the imbalanced node classification problem in HINs by generating synthetic nodes and obtaining high-quality embeddings while preserving the minority semantics. 
To this end, we propose a novel data augmentation method for HINs that incorporates the influence on minority nodes to determine the most informative neighbor nodes. Specifically, the HIN is decomposed into multiple bipartite networks according to meta-paths and the Personalised PageRank is utilized to measure the influences among different node types, thereby the most important neighbor nodes of minorities are identified to construct the networks for synthetic nodes. 
To suppress the potential noise introduced during the generation procedure, we designed two regularization approaches for HGNNs that leverage inter-type and intra-type information to constrain the embeddings of synthetic nodes. Inter-type information is derived from topological structures and is reflected in the heterogeneous links and neighbors for different types, which can enhance the node semantic representation. Intra-type information is distinguished by classes, and synthetic nodes should be constrained by class relationships. By incorporating these powerful supervision signals, our proposed method can effectively counteract the imbalance in HINs by enabling more expressive and robust node embeddings.

The main contributions of this paper are summarized as:
\begin{itemize}[leftmargin=*]
  \item \textbf{New problem and insights.} We explore an important yet under-explored problem: imbalanced node classification in HINs, and we are the first (to the best of our knowledge) to point out the challenge posed by the unique and intrinsic \textit{semantic imbalance} issue, where existing imbalanced node classification methods for homogeneous graphs cannot handle the complicated heterogeneity and rich semantics in HINs.
  
  \item \textbf{New Methodology.} We propose the first method for addressing the semantic imbalance issue in imbalanced HINs by synthesizing the target nodes and selecting neighbor nodes following minority semantics. Additionally, we design two regularization approaches to constrain the representation of synthetic nodes from both inter-type and intra-type information perspectives, thereby increasing robustness to possible noises in generated synthetic nodes and mitigating the node imbalance problem.

  \item \textbf{SOTA Performance.} Extensive experiments are conducted on four public datasets with different scales and characteristics, and the results show that our proposed method outperforms various state-of-the-art (SOTA) methods in both graph mining and general machine learning literature, including cost-sensitive learning methods, re-balanced training strategies, and data augmentation methods. 
\end{itemize}

\section{PRELIMINARIES}
\label{preliminary}
In this section, we first define some key concepts used throughout the paper, and then mathematically formulate our research problem.

\noindent\textbf{Definition 1: Heterogeneous Information Networks}.
A heterogeneous information network (HIN) is defined as $\mathcal{G}$ = ($\mathcal{V}$, $\mathcal{E}$, $\phi$, $\psi$), where $\mathcal{V}$ is the set of nodes, $\mathcal{E}$ is the set of edges. $\phi:\mathcal{V}\to \mathcal{A}$ and $\psi:\mathcal{E}\to \mathcal{R}$ are node type mapping function and edge type mapping function, respectively. $\mathcal{A}$ indicates the set of node types, and $\mathcal{R}$ denotes the set of relations, where $\left |\mathcal{A} \right | +\left |\mathcal{R} \right | > 2$.

\noindent\textbf{Definition 2: Network Schema}. 
The network schema for $\mathcal{G}$ is a meta template graph defined over $\mathcal{A}$, with edges as relations from $\mathcal{R}$, denoted as $\mathcal{T}=(\mathcal{A},\mathcal{R})$.

\noindent\textbf{Definition 3: Meta-path}. 
A meta-path $\mathcal{P}$ is a path defined on network schema and denoted in the form of ${A_1} \stackrel{R_1}{\longrightarrow} {A_2} \stackrel{R_2}{\longrightarrow} ... \stackrel{R_l}{\longrightarrow} {A_{l+1}}$, where $A_{i}\in \mathcal{A}$. It describes a composite relation $R = R_1 \circ R_2 \circ ... \circ R_l$ between node types $A_1$ and $A_{l+1}$, where $R_{i}\in \mathcal{R}$, and $\circ$ denotes the composition operator on relations.

HINs provide both the topology structure and high-level semantics of the nodes. For the HIN whose network schema is Fig. \ref{fig_pre}(a), $\left |\mathcal{A} \right |=4$ and $\left |\mathcal{R} \right |=6$. For the target node ${A}_t=$``Paper'', there are three neighbor node types for the target node, i.e., ``Term'', ``Author'', and ``Conference''. The meta-path ``Paper-Term-Paper-Author'' is one of the meta-path between ``Paper'' and ``Author'', which describes the relation between papers and the authors of related work.

\noindent\textbf{Definition 4: Heterogeneous Information Network Embedding.}
The goal of heterogeneous information network embedding is to learn a mapping function to project each node to a low-dimensional space $\mathbb{R}^{d}$, where $d \ll \left |\mathcal{V} \right |$. Node representations are able to preserve the rich structure and semantic information and can be applied to downstream tasks.
 
\noindent\textbf{Problem: Imbalanced Node Classification in HINs.} 
For multi-class classification problems in HINs, the target of classification is the target node type ${A_t} \in \mathcal{A}$, and each node of the target node type $v \in \mathcal{V}^{{A_t}}$ \footnote{The superscript of variables used in the paper indicates the type of the node or relation.} belongs to a class $\mathcal{C}_i \in \{\mathcal{C}_1, ..., \mathcal{C}_m\}$, where $m$ is number of classes, and the one-hot label can be defined as $\mathbf{Y} \in \mathbb{R}^{m}$. The distribution of class can be evaluated by the \textit{imbalance ratio}, which is $\frac{\min_i(\left |\mathcal{C}_i\right | )}{\max_i(\left |\mathcal{C}_i\right | )}$, where  $\left |\mathcal{C}_i\right | $ is the size of $i$-th class. In the imbalanced setting, the imbalance ratio of classes is small and HGNN is trained on both $\mathcal{G}$ and imbalanced $\mathbf{Y}$ to classify target nodes $v \in \mathcal{V}^{{A_t}}$ into their correct classes.

\begin{figure*}[t]
\setlength{\abovecaptionskip}{0.2cm}
\centering
\includegraphics[width=0.91\linewidth]{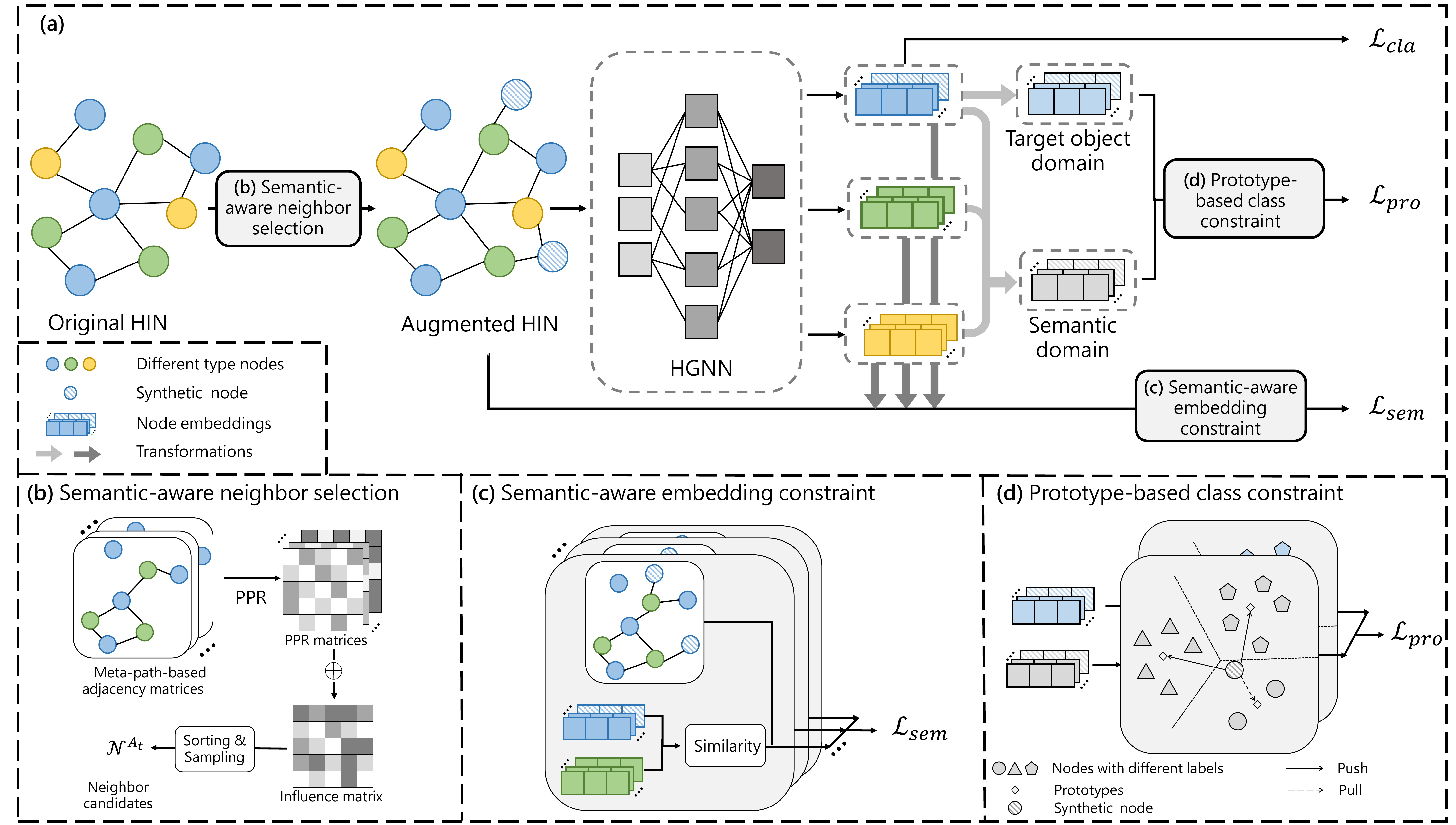}
\caption{(a) The overall framework of SNS: The original imbalanced HIN is augmented by generating the synthetic nodes, and their neighbors are selected by the semantic-aware neighbor selection. The augmented HIN is encoded by HGNN to generate the embeddings for each kind of node. (b) HIN is sampled according to various meta-paths, and the Personalized PageRank (PPR) is utilized to calculate the influence matrix. The neighbor candidate set of the synthetic node is derived by sorting and sampling processes referring to the influence matrix. (c) The semantic-aware embedding constraint is constructed according to the node embeddings and first-order network structure. (d) Synthetic node embeddings in both the target node domain and semantic domain are constrained using majority and minority class information.}
\label{fig_main}
\end{figure*}

\section{Related Work}
\subsection{Heterogeneous Information Network Embedding}
To tackle the challenge of heterogeneity, HIN embedding methods mainly follow two research lines.
Firstly, proximity-preserving methods \cite{dong2017metapath2vec, fu2017hin2vec, zhang2019shne, shi2018aspem,shi2018easing, 10.1145/3219819.3219986} leverage the random walk or graph decomposition to model the node context regarding heterogeneous semantics. For example, metapath2vec \cite{dong2017metapath2vec} utilizes the meta-paths to extract the semantic information and generate the embeddings by feeding the node sequences to skip-gram. HHNE \cite{wang2019hyperbolic} further extends metapath2vec to the hyperbolic space and PTE \cite{tang2015pte} proposes to decompose a HIN into multiple bipartite networks. 
Unlike these shallow network embedding methods, message-passing methods adopt the aggregation/message-passing process to propagate the information among neighbor nodes. 
Meta-path-based models \cite{fu2020magnn, schlichtkrull2018modeling, wang2019heterogeneous, zhang2019heterogeneous}, such as HAN \cite{wang2019heterogeneous}, propagate messages at the scope of each meta-path to generate semantic vectors and then aggregate them across meta-paths using semantic aggregation. However, these methods rely heavily on domain knowledge to extract meta-paths, and separating different semantics using human priors can impair the implicit topology information within the HIN.
To avoid potential information loss, meta-path-free methods \cite{zhu2019relation, hong2020attention, hu2020heterogeneous, lv2021we, yang2022simple, yang2021interpretable} leverage a hierarchical structure to capture structural and semantic information simultaneously. These methods propagate semantic information among various node types across different kinds of edges. For example, HGT \cite{hu2020heterogeneous} proposes to use each edge's type to parameterize the Transformer-like self-attention architecture and perform aggregation for all kinds of neighbors. On the other hand, ie-HGCN \cite{yang2021interpretable} employs node-level and type-level aggregation to automatically discover and exploit the most useful meta-paths for each target node, providing interpretable results.

\subsection{Imbalanced Node Classification in Homogeneous Graphs}
The problem of imbalanced node classification in homogeneous graphs has received significant attention from many researchers. However, these methods are designed exclusively for homogeneous graphs and disregard heterogeneous edges, which results in their inability to generalize to imbalanced HINs. GraphSMOTE \cite{zhao2021graphsmote} and ImGAGN \cite{qu2021imgagn} introduce the SMOTE \cite{chawla2002smote} and GAN \cite{goodfellow2020generative} in graph augmentation respectively and generate the synthetic minority nodes while learning the topological structure distribution at the same time. However, the link predictors in their designs can only handle homogeneous edges that connect nodes of the same type.
DR-GCN \cite{shi2020multi} and RA-GCN \cite{GHORBANI2022102272} learn higher weights for under-represented classes in an adversarial training manner, but their focus is limited to the information among the target nodes. The neglect of heterogeneous neighbors in HINs hinders the exploration of relationships among different types of nodes.
Besides, ReNode \cite{chen2021topology} and TAM \cite{song2022tam} argue that the imbalanced label location also impairs the classification performance of minority classes and mitigates this topology imbalance problem by topology-aware weight and margin loss design, respectively. These methods rely on the assumption of homophily \cite{mcpherson2001birds} and require the presence of homogeneous edges among target nodes. Unfortunately, this assumption is often violated in most HINs, thereby limiting the practicality of these approaches.
Recently, GraphENS \cite{park2022graphens} studies the over-fitting problem to the neighbor sets of minor classes and proposes an augmentation method to synthesize the whole ego network for the minority class. However, the effectiveness of GraphENS in HINs remains unexplored as it is designed within the context of homogeneous neighbors, leaving its applicability to heterogeneous neighbors uncertain and in need of further investigation.

\section{METHODOLOGIES}
This section presents SNS, the first method for semantic imbalance problem in imbalanced HINs that synthesizes both the node and semantics for minor classes by considering the class influences. To preserve the semantics of minority nodes, we first introduce the semantic-aware neighbor selection in Section \ref{S31} and then explain how to further enhance the semantic and class representation of synthetic nodes by utilizing the inter-type and intra-type information in Section \ref{S32} and \ref{S33}, respectively.

\subsection{Semantic-aware neighbor selection and attribute synthesis}
\label{S31}
Semantic-aware neighbor selection is designed to determine which neighbor node to be connected to a synthetic target node according to different relations. The goal is to expose a synthetic node to environments that conform to the minority semantics while preventing the generation of noisy connections. To this end, we construct the neighbor node distribution for each neighbor node type and stochastically sample the neighbor nodes from these distributions.
For the sake of simplicity, we take the target node type $A_t$ and one of its neighbor node types $A_k$ for illustration. First, we consider the first-order relation between $A_k$ and $A_t$, i.e., $A_k\stackrel{R_1}\longrightarrow A_t$, where $R_1 \in \mathcal{R}$. Then, we extend $R_1$ by its reversed relation and leverage the bidirectional relation $\bar{R}_1$ (e.g., ``Author-writes-Paper'' and ``Paper-is written by-Author'') to construct a symmetrical adjacency matrix $\mathbf{A}^{\bar{R}_1}$, where $\mathbf{A}^{\bar{R}_1} \in \mathbb{R}^{({\left |{A}_k\right |+\left |{A}_t\right |})\times({\left |{A}_k\right |+\left |{A}_t\right |})}$. 
With the subgraph represented by the symmetrical adjacency matrix, we can measure the influence of each neighbor node on the target node type according to the topology structure. Here, we use the well-known Personalized PageRank $\Pi^{\bar{R}_1}$ \cite{page1999pagerank} as it can directly incorporate multi-hop neighborhood information without explicit message passing: 
\begin{equation}
\label{ppr}
\Pi^{\bar{R}_1}=\alpha(I-(1-\alpha)\tilde{\mathbf{A}}^{\bar{R}_1})^{-1},
\end{equation}
where $I$ is the identity matrix, $\alpha \in(0, 1]$ is the random walk restart probability, $\tilde{\mathbf{A}}^{\bar{R}_1}=D^{-\frac{1}{2}} \mathbf{A}^{\bar{R}_1} D^{-\frac{1}{2}} $is the adjacency matrix normalized by the diagonal degree matrix $D$. A larger value of $\Pi^{\bar{R}_1}_{i,j}$, i.e., the element of $i$-th row and $j$-th column in $\Pi^{\bar{R}_1}$, indicates that node $v_i$ has more influence on another node $v_j$ in the subgraph $\mathbf{A}^{\bar{R}_1}$. 
Note that Eq. (\ref{ppr}) can be efficiently computed using the power-iteration method or approximation techniques \cite{bojchevski2020pprgo}, facilitating the scalability for large-scale HINs.
To comprehensively capture the influence of neighbor nodes, we optionally take into account higher-order relations captured by meta-paths between $A_k$ and $A_t$ in the HIN, and the total influence between $A_k$ and $A_t$ can be measured by Eq. (\ref{inf}):
\begin{equation}
\label{inf}
\Pi^{A_k}=\sum_{c=1}^{n}\Pi^{\bar{R}_c},
\end{equation}
where $n$ is the number of meta-paths between $A_k$ and $A_t$.
Notice that the first-order relation between $A_k$ and $A_t$ refers to the 1-hop neighbors in the HIN (e.g., ``Author-writes-Paper''), while higher-order relations represent multi-hop neighbors (e.g., ``Author-Paper-Term-Paper'', which indicates the author of the related paper). These higher-order relations extract richer semantics between $A_k$ and $A_t$ and can alleviate the potential influence sparsity problem, which may arise due to the presence of disconnected components in a single subgraph $\mathbf{A}^{\bar{R}_c}$.

Then, the influence of the neighbor node $v_j$ on the target minority class $\mathcal{C}_l$ is measured by Eq. (\ref{ninf}):
\begin{equation}
\label{ninf}
P_j=\sum_{v_i\in \mathcal{C}_l} \Pi^{A_k}_{i,j}.
\end{equation}

The influence of neighbor nodes in type $A_k$ on the target minority $\mathcal{C}_l$ is $P \in \mathbb{R}^{1\times{\left |{A}_k\right | }}$, and top-$k$ neighbor nodes with the highest influence scores are selected as neighbor candidates $\mathcal{N}^{A_{k}}$ for synthetic nodes. $k$ in top-$k$ is calculated by $\mu \times d_{max}^{A_k}$, where $\mu$ is a hyper-parameter controlling the candidate size and $d_{max}^{A_k}$ is the maximum degree of minority nodes related to the neighbor type $A_k$. Utilizing $d_{max}^{A_k}$ allows for the adaptive adjustment of candidate sizes for different neighbor types based on their respective minority degree distributions. As $\mu$ increases, the candidate size grows and more neighbor nodes are added in $\mathcal{N}^{A_{k}}$ until all neighbor nodes are included.

Given the neighbor candidates $\mathcal{N}^{A_{k}}$, we sample the neighbors of synthetic nodes from $\mathcal{N}^{A_{k}}$ randomly. The number of neighbors is sampled from the degree distribution of minority nodes to resemble the basic structural patterns of the original HIN.

Aside from the network structures, target nodes in some HINs also contain rich attributes. To generate the attributes for a synthetic node, we randomly select two nodes from minority classes and interpolate their attributes. However, the plain interpolation may produce attributes that are out of distribution and harm the classification performance. Inspired by \cite{park2022graphens}, we retain the $K\%$ most important attributes of one node and calculate the remaining attributes by the linear interpolation to avoid the over-fitting problem. Specifically, the importance of the $j$-th attribute of the node $v_i$ is calculated as the gradient magnitude w.r.t. the $j$-th attribute as Eq. (\ref{attri}):
\begin{equation}
\label{attri}
s_j=\left |\frac{\partial \mathcal{L}}{\partial X_i} \right | _j,
\end{equation}
where $\mathcal{L}$ is the loss for model training and $X_i$ is the attributes of the node $v_i$. $s_j$ is derived from the previous training iteration and the $K\%$ attributes with the largest $s_j$ are preserved in the attribute of the generated synthetic node.

\subsection{Semantic-aware Embedding Constraint}
\label{S32}
The semantic information of the target node is represented by various neighbor node types.
Therefore, to further enhance the semantics of embeddings for synthetic nodes generated by HGNN and suppress potential noise in the synthetic node generation process, SNS utilizes the inter-type information and employs the embeddings of various types of neighbor nodes to enforce constraints during the training of synthetic nodes, as illustrated in Fig. \ref{fig_main}(c). 

After the node augmentation in Section \ref{S31}, the target node set ${\mathcal{V}}^{A_{t}}$ is expanded to $\tilde{\mathcal{V}}^{A_{t}}$ by synthetic node set $\acute{\mathcal{V}}^{A_{t}}$, i.e., $\tilde{\mathcal{V}}^{A_{t}}={\mathcal{V}}^{A_{t}}\cup \acute{\mathcal{V}}^{A_{t}}$. Then, we make use of neighbor nodes of synthetic nodes within the first-order relation $A_k\stackrel{R_1}\longrightarrow A_t$, i.e., 1-hop neighbor nodes of synthetic nodes $\hat{\mathcal{V}}^{A_{k}} \in \mathcal{V}^{A_{k}}$, to strengthen the constraints on synthetic nodes by considering their most influential association. The semantic loss for neighbor type $A_k$ is quantified based on the neighbor node set $\hat{\mathcal{V}}^{A_{k}}$ and expanded target node set $\tilde{\mathcal{V}}^{A_{t}}$ by Eq. (\ref{linkloss}):
\begin{equation}
\begin{aligned}
\small
&\mathcal{L}_{sem}^{A_k}=-\frac{1}{\left | {\tilde{\mathcal{V}}^{A_{t}}} \right |} \sum_{v_i\in {\tilde{\mathcal{V}}^{A_{t}}}}(\frac{1}{ \left | {\hat{\mathcal{V}}^{A_{k}}} \right | } \sum_{v_j\in {\hat{\mathcal{V}}^{A_{k}}}}\mathbf{A}^{R_1}_{i,j}\log\sigma(\boldsymbol{p}_i\boldsymbol{p}_j^{\mathsf{T}})),\\
& \ \ \  \ \ \ \ \  \ \ \ \ \ \  \boldsymbol{p}_i = {\rm{MLP}}_s^{A_{t}}(\boldsymbol{z}_i), \ \ \ \ \ \ \  \ \boldsymbol{p}_j = {\rm{MLP}}_s^{A_{k}}(\boldsymbol{z}_j), \\
\end{aligned} 
\label{linkloss}
\end{equation}
where $\mathbf{A}^{R_1}_{i,j}$ is the element of $i$-th row and $j$-th column in the adjacency matrix $\mathbf{A}^{R_1}$ and $\sigma$ is the sigmoid function. Node embeddings $\boldsymbol{z}_i$ and $\boldsymbol{z}_j$ are generated by HGNN for nodes $v_i$ and $v_j$, respectively. ${\rm{MLP}}_s^{A_{t}}$ and ${\rm{MLP}}_s^{A_{k}}$ are type-specific multi-layer perceptrons for ${A_{t}}$ and ${A_{k}}$, which projects the embeddings into the subspace to measure similarities. Notice that $\tilde{\mathcal{V}}^{A_{t}}$ contains both real target nodes and synthetic nodes, and $\hat{\mathcal{V}}^{A_{k}}$ only includes the 1-hop neighbor nodes of synthetic nodes. This allows $\mathcal{L}_{sem}^{A_k}$ to promote semantic consistency between the neighbors of synthetic nodes ${\hat{\mathcal{V}}^{A_{k}}}$ and real target nodes ${\mathcal{V}}^{A_{t}}$, and to subsequently constrain the representation of synthetic nodes $\acute{\mathcal{V}}^{A_{t}}$ by referring to the enhanced neighbor nodes ${\hat{\mathcal{V}}^{A_{k}}}$. It is worth mentioning that the semantic loss $\mathcal{L}_{sem}^{A_k}$ can also be regarded as the link prediction loss for relation $R_1$, which is only applicable to ${\hat{\mathcal{V}}^{A_{k}}}$. 

Finally, the overall semantic loss is calculated via Eq. (\ref{linkloss2}):
\begin{equation}
\begin{aligned}
\small
&\mathcal{L}_{sem}=\sum_{A_k\in {\mathcal{\tilde{A}}}}\mathcal{L}_{sem}^{A_k},\\
\end{aligned} 
\label{linkloss2}
\end{equation}
where $\mathcal{\tilde{A}} \in \mathcal{{A}}$ is the neighbor node type set which only includes the first-order relations of the target node type.

\begin{table*}[t]
\setlength{\abovecaptionskip}{0.2cm}
\caption{{Properties of four datasets.}}
\centering
 \label{tab_dataset}
 \resizebox{0.8\textwidth}{!}{
\begin{tabular}{l|lllllll}
\toprule
Datasets & \#Nodes    & Node type (\#)                         & \makecell[c]{Target \\node} & Relations                                                                                        & \# Classes & \makecell[c]{\# Minority \\classes} & \makecell[c]{\# Nodes of  \\ majority classes}\\ \hline
MAG & 26334    & P (4017), A (15383), I (1480), F (5454) & P             & P$\rightleftharpoons$P, P$\rightleftharpoons$F, P$\rightleftharpoons$A,   A$\rightleftharpoons$I & 4          & 2     & 60              \\
ACM & 11252    & P (4025), A (7167), S (60)              & P             & P$\rightleftharpoons$A, P$\rightleftharpoons$S                                                   & 3          & 1         & 80          \\
DBLP & 27303   & A (4057), P (14328), C (20), T (8898)   & A             & P$\rightleftharpoons$A, P$\rightleftharpoons$C, P$\rightleftharpoons$T                           & 4          & 2     & 60              \\
IMDB & 50000   & M (3328), A (42553), U (2103), D (2016) & M             & M$\rightleftharpoons$A, M$\rightleftharpoons$U, M$\rightleftharpoons$D                           & 4          & 2     & 40              \\ \bottomrule
\end{tabular}}
\end{table*}

\subsection{Prototype-based Class Constraint}
\label{S33}

In addition to leveraging semantic information across multiple node types, the intra-type information provides crucial class-related insights. To comprehensively explore class relations and enforce class constraints, SNS employs prototype-based class constraints on both the synthetic embeddings and their neighbor embeddings, as depicted in Fig. \ref{fig_main}(d).

Firstly, all node embeddings $\boldsymbol{z}^{A_k}_i$ ($\forall v_i\in {\mathcal{V}}^{A_{k}}$ and $\forall {A_{k}} \in  \mathcal{A}$) are projected into the subspace as Eq. (\ref{mlp}):
\begin{equation}
\boldsymbol{q}_i^{A_k} = {\rm{MLP}}_p^{A_{k}}(\boldsymbol{z}_i^{A_k}),
\label{mlp}
\end{equation}
where ${\rm{MLP}}_p^{A_{k}}$ is type-specific multi-layer perceptrons for ${A_{k}}$.
Then, to constrain the node embedding and its semantics simultaneously, a new semantic domain is constructed for each target node $v_i\in {\tilde{\mathcal{V}}^{A_{t}}}$ according to the network schema. The semantic embedding $\boldsymbol{g}_i^{A_t}$ for the node $v_i$ is calculated by Eq. (\ref{semdomain}):

\begin{equation}
\boldsymbol{g}_i^{A_t}=\frac{1}{1+\left | \mathcal{\tilde{A}} \right | } (\boldsymbol{q}_i^{A_t}+\sum_{A_k\in  \mathcal{\tilde{A}}} (\frac{1}{\left | {N^{A_k}} \right |}\sum_{v_j\in{N^{A_k}}}\boldsymbol{q}_i^{A_k})),
\label{semdomain}
\end{equation}
where $N^{A_k}$ is the neighbor node set of target node $v_i$ in neighbor type $A_k$, and $\mathcal{\tilde{A}} \in \mathcal{{A}}$ is the neighbor node type set. Each semantic embedding is constructed by two processes in Eq. (\ref{semdomain}), which include intra-type aggregation and inter-type aggregation.
Subsequently, the class prototype (i.e., class centroid) embeddings \cite{snell2017prototypical} are generated in the target node domain and the semantic domain by Eq. (\ref{obj_cent}) and (\ref{sem_cent}), respectively:

\begin{equation}
\boldsymbol{e}_j=\frac{1}{\left | \mathcal{C}_j \right | } \sum_{v_i\in \mathcal{C}_j} \boldsymbol{q}_i^{A_t},
\label{obj_cent}
\end{equation}

\begin{equation}
\boldsymbol{o}_j=\frac{1}{\left | \mathcal{C}_j \right | } \sum_{v_i\in \mathcal{C}_j} \boldsymbol{g}_i^{A_t},
\label{sem_cent}
\end{equation}
where $\mathcal{C}_j \in \{\mathcal{C}_1, ..., \mathcal{C}_m\}$ is the class set for real target nodes, and $m$ is number of classes.
Finally, the class constraints are defined for synthetic nodes in both two domains to encourage the consistency between the embeddings of synthetic nodes and their minority prototype and push them away from the other prototypes:
\begin{equation}
\begin{aligned}
&  \ \ \ \ \ \ \ \ \ \ \ \ \ \ \ \ \ \ \ \ \ \ \ \ \ \ \ \ \ \ \mathcal{L}_{pro}=\frac{\mathcal{L}_e+\mathcal{L}_o}{2},\\
&\mathcal{L}_e=-\frac{1}{\left | {{{N}}_{c}} \right |} \sum_{c\in {{N}}_{c}} (\frac{1}{ \left | {{\mathcal{\acute{V}}}_c^{A_{t}}} \right | } \sum_{v_i\in {{\mathcal{\acute{V}}}_c^{A_{t}}}}\log\frac{\exp(\boldsymbol{q}^{A_t}_i\boldsymbol{e}_c^{\mathsf{T}}/T)}{ {\textstyle \sum_{j=1}^{m}\exp(\boldsymbol{q}^{A_t}_i\boldsymbol{e}_j^{\mathsf{T}}/T)} } ),\\
&\mathcal{L}_o=-\frac{1}{\left | {{{N}}_{c}} \right |} \sum_{c\in {{N}}_{c}} (\frac{1}{ \left | {{\mathcal{\acute{V}}}_c^{A_{t}}} \right | } \sum_{v_i\in {{\mathcal{\acute{V}}}_c^{A_{t}}}}\log\frac{\exp(\boldsymbol{g}^{A_t}_i\boldsymbol{o}_c^{\mathsf{T}}/T)}{ {\textstyle \sum_{j=1}^{m}\exp(\boldsymbol{g}^{A_t}_i\boldsymbol{o}_j^{\mathsf{T}}/T)} } ),\\
\end{aligned}
\label{cla_loss}
\end{equation}
where ${N}_{c}$ is the minority class set and ${\mathcal{\acute{V}}}_c^{A_{t}}$ is the target synthetic node set of class $c$. $T$ is the temperature that controls the strength of penalties.

In the end, with the definition of $\mathcal{L}_{sem}$ and $\mathcal{L}_{pro}$, the HGNN is trained together with the Cross-Entropy loss $\mathcal{L}_{cla}$ as shown in Eq. (\ref{totoalloss}):
\begin{equation}
\begin{aligned}
\mathcal{L}=\mathcal{L}_{cla}+\lambda_1\mathcal{L}_{sem}+\lambda_2\mathcal{L}_{pro},\\
\end{aligned}
\label{totoalloss}
\end{equation}
where $\lambda_1$ and $\lambda_2$ are two hyper-parameters to balance the losses, and $\mathcal{L}_{sem}$ and $\mathcal{L}_{pro}$ are designed exclusively for synthetic nodes.

\section{Experiments}
We test SNS on real-world HINs with different properties to verify the effectiveness and aim to answer the following questions. 

\noindent\textbf{Q1}: Compared to other methods designed for imbalanced classification problems, particularly state-of-the-art (SOTA) method in homogeneous graphs, can SNS achieve better classification performance?
\noindent\textbf{Q2}: Does Semantic-aware neighbor selection in SNS outperform other neighbor selection methods? 
\noindent\textbf{Q3}: How does each component in SNS affect the model performance? 
\noindent\textbf{Q4}: How about the performance of SNS under different imbalance ratios? 
\noindent\textbf{Q5}: Can SNS generalize well to different HGNN models? 
\noindent\textbf{Q6}: How do hyper-parameters affect SNS? 

\begin{table*}[pt]
\setlength{\abovecaptionskip}{0.2cm}
\caption{{Performance of different baselines with ie-HGCN as the base model. The best performance is bold.}}
 \label{tab_end2end}
 \resizebox{\textwidth}{!}{
\begin{tabular}{l|lll|lll|lll|lll}
\toprule
                   & \multicolumn{3}{c|}{MAG}             & \multicolumn{3}{c|}{ACM}             & \multicolumn{3}{c|}{DBLP}            & \multicolumn{3}{c}{IMDB}             \\ \hline
Method             & F1 score   & ACC        & BACC       & F1 score   & ACC        & BACC       & F1 score   & ACC        & BACC       & F1 score   & ACC        & BACC       \\ \hline
Vanilla            & 83.02$\pm$3.69 & 83.55$\pm$3.42 & 82.64$\pm$3.68 & 59.31$\pm$0.77 & 76.16$\pm$3.26 & 61.13$\pm$1.03 & 89.32$\pm$1.02 & 89.90$\pm$0.89  & 89.66$\pm$0.96 & 27.38$\pm$1.41 & 33.17$\pm$0.90  & 40.44$\pm$0.85 \\
Reweight           & 89.68$\pm$3.99 & 89.98$\pm$3.52 & 89.65$\pm$3.69 & 59.93$\pm$2.00  & 73.61$\pm$1.49 & 61.22$\pm$1.62 & 89.95$\pm$0.82 & 90.69$\pm$0.74 & 90.09$\pm$0.79 & 32.77$\pm$1.49 & 35.87$\pm$1.22 & 43.29$\pm$2.00  \\
Focal loss         & 91.64$\pm$3.82 & 91.77$\pm$3.62 & 91.59$\pm$3.78 & 61.16$\pm$2.11 & 72.98$\pm$2.35 & 62.50$\pm$1.64  & 91.54$\pm$0.38 & 92.10$\pm$0.37  & 91.58$\pm$0.39 & 32.35$\pm$1.76 & 36.66$\pm$2.02 & 41.93$\pm$1.65 \\
BS & 93.04$\pm$2.19 & 93.07$\pm$2.16 & 93.08$\pm$2.13 & 59.59$\pm$0.99 & 75.30$\pm$3.06  & 60.71$\pm$0.60  & 89.35$\pm$1.76 & 89.85$\pm$1.76 & 89.59$\pm$1.54 & 33.44$\pm$1.53 & 37.99$\pm$5.05 & 40.16$\pm$2.90  \\
Self-training      & 93.51$\pm$1.38 & 93.53$\pm$1.33 & 93.34$\pm$1.36 & 62.06$\pm$2.97 & 76.61$\pm$2.03 & 63.52$\pm$1.58 & 89.34$\pm$1.57 & 89.99$\pm$1.45 & 89.86$\pm$1.38 & 33.44$\pm$3.49 & 38.20$\pm$2.17  & 43.06$\pm$2.92 \\
cRT                & 90.36$\pm$1.42 & 90.38$\pm$1.44 & 89.95$\pm$1.53 & 60.55$\pm$0.80  & 74.06$\pm$2.06 & 60.55$\pm$0.59 & 89.87$\pm$0.62 & 90.39$\pm$0.67 & 89.77$\pm$0.69 & 29.83$\pm$1.45 & 32.84$\pm$2.32 & 42.96$\pm$0.70  \\
ROS                & 85.88$\pm$3.64 & 86.18$\pm$3.35 & 85.72$\pm$3.43 & 60.17$\pm$1.71 & 74.52$\pm$2.31 & 61.20$\pm$1.52  & 90.28$\pm$0.87 & 90.91$\pm$0.92 & 90.36$\pm$0.78 & 32.32$\pm$3.23 & 36.28$\pm$3.62 & 42.37$\pm$2.03 \\
SMOTE              & 84.16$\pm$3.89 & 84.96$\pm$3.55 & 84.24$\pm$3.76 & 59.41$\pm$1.12 & 76.80$\pm$2.30   & 61.55$\pm$0.79 & 89.70$\pm$0.84  & 90.33$\pm$0.78 & 90.09$\pm$0.57 & 28.32$\pm$0.89 & 33.06$\pm$1.78 & 40.46$\pm$1.59 \\
GraphENS           & 93.30$\pm$2.39  & 93.33$\pm$2.39 & 93.35$\pm$2.28 & 61.52$\pm$0.67 & 72.45$\pm$2.85 & 61.45$\pm$0.57 & 91.62$\pm$0.19 & 92.17$\pm$0.22 & 91.39$\pm$0.22 & 33.80$\pm$2.71  & 37.56$\pm$2.03 & 42.97$\pm$4.05 \\
SNS           & \textbf{95.72$\pm$0.77} & \textbf{95.68$\pm$0.77} & \textbf{95.62$\pm$0.81} & \textbf{63.84$\pm$0.78} & \textbf{77.31$\pm$2.56} & \textbf{64.25$\pm$0.59} & \textbf{92.50$\pm$0.53}  & \textbf{93.01$\pm$0.47} & \textbf{92.48$\pm$0.60}  & \textbf{35.13$\pm$1.36} & \textbf{39.41$\pm$2.47} & \textbf{43.73$\pm$1.76} \\ \bottomrule
\end{tabular}}
\end{table*}

\subsection{Experimental Settings}

\noindent\textbf{Baselines}. 
We compare our proposed method with 8 state-of-the-art methods designed for the class imbalance problem in 3 categories, which include: cost-sensitive learning methods, re-balanced training strategies, and data augmentation methods. {Reweight}~\cite{ren2018learning}: a cost-sensitive method by increasing the category weight of classification loss function; {Focal loss}~\cite{lin2017focal}: apply a modulating term to the cross entropy loss in order to focus learning on hard misclassified examples; {Balanced softmax (BS)}~\cite{ren2020balanced}: accommodate the imbalanced label distribution shift by using an unbiased extension of Softmax; {Self-training}~\cite{li2018deeper}: a semi-supervised method by labeling the pseudo labels for unlabeled nodes and iterative optimizing the classifier; {Classifier Re-training (cRT)}~\cite{kang2019decoupling}: decouple the feature representation and classifier training procedure to avoid the biased classifier; {Random over-sampling (ROS)}: sample the nodes and their edges in minority classes to re-balance the classes; {SMOTE}~\cite{chawla2002smote}: interpolate a minority embedding and its nearest neighbor embeddings in the same class; {GraphENS}~\cite{park2022graphens}: synthesize feasible ego networks based on the similarity between source ego networks, and block the injection of harmful features in generating the mixed nodes using node feature saliency.

\noindent\textbf{Datasets}. 
We evaluate our proposed method on four publicly available HIN benchmark datasets with different scales and characteristics, which include: MAG, ACM, DBLP and IMDB. {MAG} is a subset of Microsoft Academic Graph. It contains four node types: Paper (P), Author (A), Institution (I), and Field (F), and eight relations among them. 
Paper nodes are labeled into four classes according to their published venues: IEEE Journal of Photovoltaics, Astrophysics, Low Temperature Physics, and Journal of Applied Meteorology and Climatology. 
{ACM} is extracted from ACM digital library. It contains three node types: Paper (P), Author (A), and Subject (S), and four relations among them. 
Paper nodes are divided into three classes: Data Mining, Database, and Computer Network. 
{DBLP} is extracted from DBLP bibliography. It contains four node types: Author (A), Paper (P), Conference (C), and Term (T), and six relations among them. Author (A) nodes are labeled according to their four research areas: Data Mining, Information Retrieval, Database, and Artificial Intelligence. 
{IMDB} is extracted from the online movie rating website IMDB. It contains four node types: Movie (M), Actor (A), User (U) and Director (D), and six relations among them. Movie (M) nodes are categorized into four classes according to their genres: Comedy, Documentary, Drama, and Horror.
We adopt the same basic dataset and imbalance setting as previous works: \cite{yang2022self} and \cite{park2022graphens}, where the label rate is fixed at 6\% for all datasets.
The detailed dataset statistics are summarized in Table \ref{tab_dataset}. If not specified otherwise, $imbalance\ ratio$ is set to 0.1 for all datasets and experiments.

\noindent\textbf{Evaluation metrics}. The performance of each baseline is evaluated by three well-established evaluation metrics, as used in \cite{park2022graphens}: accuracy (ACC), F1 score, and balanced accuracy (BACC). ACC is the ratio of corrected samples among test samples. The F1 score is the harmonic mean of precision and recall for each class. BACC is defined as the average of recall obtained in each class. Note that, the F1 score and BACC are designed for imbalanced data, avoiding the dominance of the majority classes in the final performance.

\begin{table*}[pt]
\setlength{\abovecaptionskip}{0.2cm}
\caption{{Ablation study results on four datasets. The best performance is bold.}}
 \label{tab_abl}
 \resizebox{\textwidth}{!}{
\begin{tabular}{l|lll|lll|lll|lll}
\toprule
                                   & \multicolumn{3}{c|}{MAG}             & \multicolumn{3}{c|}{ACM}             & \multicolumn{3}{c|}{DBLP}            & \multicolumn{3}{c}{IMDB}             \\ \hline
Method                             & F1 score   & ACC        & BACC       & F1 score   & ACC        & BACC       & F1 score   & ACC        & BACC       & F1 score   & ACC        & BACC       \\ \hline
SNS w/o NS (m) & 89.99$\pm$1.72 & 91.15±1.70 & 91.71$\pm$1.81 & 61.34$\pm$2.21 & 75.20$\pm$5.05 & 63.04$\pm$0.93 & 91.02$\pm$0.49 & 91.60$\pm$0.44  & 91.17$\pm$0.56 & 34.44$\pm$2.62 & 38.31$\pm$3.21 & 42.10$\pm$1.60 \\
SNS w/o NS (a)              & 91.62$\pm$2.85 & 91.79$\pm$2.61 & 91.68$\pm$2.76 & 60.31$\pm$1.50 & 73.37$\pm$5.57 & 61.30$\pm$1.53 & 89.16$\pm$0.57 & 89.72$\pm$0.51 & 88.72$\pm$0.61 & 33.15$\pm$4.64 & 38.02$\pm$5.38 & 41.28$\pm$2.12 \\

SNS w/o PC     & 95.26$\pm$0.58 & 95.22$\pm$0.58 & 95.15$\pm$0.63 & 61.87$\pm$3.55 & 73.05$\pm$5.10 & 61.99$\pm$3.74 & 92.46$\pm$0.65 & 92.98$\pm$0.61 & 92.31$\pm$0.67 & 34.53$\pm$2.80 & 38.52$\pm$3.98 & 43.27$\pm$1.54 \\
SNS w/o SE     & 94.89$\pm$0.39 & 94.84$\pm$0.39 & 94.73$\pm$0.40 & 61.72$\pm$0.49 & 72.76$\pm$2.56 & 61.57$\pm$0.74 & 92.45$\pm$0.50 & 92.96$\pm$0.45 & 92.44$\pm$0.58 & 33.93$\pm$1.53 & 37.99$\pm$2.05 & 43.15$\pm$0.49 \\

SNS            &  \textbf{95.72$\pm$0.77} & \textbf{95.68$\pm$0.77 } & \textbf{95.62$\pm$0.81 } & \textbf{63.84$\pm$0.78 } & \textbf{77.31$\pm$2.56 } & \textbf{64.25$\pm$0.59 } & \textbf{92.50$\pm$0.53  } & \textbf{93.01$\pm$0.47 } & \textbf{92.48$\pm$0.60 } & \textbf{35.13$\pm$1.36 } & \textbf{39.41$\pm$2.47 } & \textbf{43.73$\pm$1.76} \\ \bottomrule
\end{tabular}}
\end{table*}

\noindent\textbf{Implementation and Settings}. 
For a fair comparison, all baselines are tested on the same base HGNN model, ie-HGCN \cite{yang2021interpretable}, unless otherwise stated.
The hyper-parameters used in experiments follow the base model or are searched by the grid search method, and we use the ADAM optimization algorithm to train all the models. Specifically, the values for $\mu$ and $T$ are searched from \{1, 3, 5, 10, 30, 50, 100, ALL\} and \{0.1, 0.5, 1, 2, 5, 10\}, respectively. ALL indicates all neighbor type nodes are selected as neighbor candidates. $\lambda_1$ and $\lambda_2$ are searched from \{0.01, 0.1, 0.5, 0.7, 1, 1.5, 2\}. The proportion of retained attributes is $K\%=10\%$. 
To eliminate randomness, we repeat each experiment 5 times and report the average test score and standard deviation.
The codes are written in Python 3.9 and the operating system is Ubuntu 16.0. We use Pytorch 1.11.0 on CUDA 11.7 to train models on GPU. All experiments are conducted on a machine with Intel(R) Xeon(R) CPUs (Gold 5120 @ 2.20GHz) and NVIDIA TITAN RTX GPUs with 24GB GPU memory. 
 
\begin{table*}[t]
\setlength{\abovecaptionskip}{0.2cm}
\centering
\caption{{Experiment results of different baselines on MAG under various imbalance ratios. The best performance is bold.}}
 \label{tab_ibr}
 \resizebox{\textwidth}{!}{
\begin{tabular}{l|lll|lll|lll|lll}
\hline
              & \multicolumn{3}{c|}{0.2}                                        & \multicolumn{3}{c|}{0.3}                                        & \multicolumn{3}{c|}{0.4}                                        & \multicolumn{3}{c}{0.5}                                         \\ \hline
Method        & F1 score            & ACC                 & BACC                & F1 score            & ACC                 & BACC                & F1 score            & ACC                 & BACC                & F1 score            & ACC                 & BACC                \\ \hline
Vanilla       & 91.15$\pm$1.81          & 91.18$\pm$1.74          & 90.96$\pm$1.84          & 94.05$\pm$0.56          & 94.10$\pm$0.60          & 93.79$\pm$0.64          & 95.54$\pm$0.75          & 95.52$\pm$0.74          & 95.50$\pm$0.78           & 96.11$\pm$0.75          & 96.08$\pm$0.74          & 96.00$\pm$0.80          \\
Reweight      & 93.51$\pm$2.29          & 93.55$\pm$2.22          & 93.43$\pm$2.38          & 94.93$\pm$1.86          & 94.97$\pm$1.79          & 94.77$\pm$1.93          & 95.75$\pm$1.18          & 95.72$\pm$1.18          & 95.74$\pm$1.20          & 96.60$\pm$0.89          & 96.57$\pm$0.90          & 96.51$\pm$0.95          \\
Focal loss    & 93.52$\pm$2.89          & 93.64$\pm$2.78          & 93.42$\pm$2.97          & 95.23$\pm$0.93          & 95.27$\pm$0.91          & 95.04$\pm$0.98          & 96.01$\pm$0.74          & 95.99$\pm$0.74          & 96.00$\pm$0.76          & 96.63$\pm$1.35          & 96.62$\pm$1.35          & 96.56$\pm$1.35          \\
BS            & 94.79$\pm$2.30          & 94.84$\pm$2.16          & 94.24$\pm$2.33          & 94.84$\pm$1.63          & 94.87$\pm$1.58          & 94.72$\pm$1.67          & 96.34$\pm$0.72          & 96.33$\pm$0.72          & 96.35$\pm$0.74          & 96.42$\pm$1.06          & 96.42$\pm$1.05          & 96.40$\pm$1.10          \\
Self-training & 94.79$\pm$1.43          & 94.82$\pm$1.36          & 94.66$\pm$1.47          & 95.06$\pm$0.88          & 95.02$\pm$0.88          & 95.03$\pm$0.95          & 96.03$\pm$0.93          & 96.10$\pm$0.93          & 96.08$\pm$0.93          & 96.16$\pm$0.46          & 96.15$\pm$0.46          & 96.07$\pm$0.49          \\
cRT           & 93.97$\pm$1.49          & 93.98$\pm$1.44          & 93.88$\pm$1.50          & 94.53$\pm$0.49          & 94.57$\pm$0.53          & 94.32$\pm$0.52          & 96.04$\pm$0.81          & 96.02$\pm$0.81          & 96.02$\pm$0.83          & 96.13$\pm$0.61          & 96.10$\pm$0.61          & 96.02$\pm$0.65          \\
ROS           & 92.23$\pm$0.85          & 92.17$\pm$0.85          & 92.16$\pm$0.91          & 94.55$\pm$0.47          & 94.53$\pm$0.49          & 94.40$\pm$0.47          & 95.96$\pm$0.42          & 95.72$\pm$0.43          & 95.67$\pm$0.42          & 95.78$\pm$1.56          & 95.74$\pm$1.58          & 95.63$\pm$1.67          \\
SMOTE         & 92.29$\pm$2.86          & 92.48$\pm$2.59          & 92.14$\pm$2.80          & 94.79$\pm$1.71          & 94.80$\pm$1.68          & 94.58$\pm$1.84          & 96.22$\pm$0.60          & 96.19$\pm$0.58          & 96.18$\pm$0.62          & 96.32$\pm$0.48          & 96.29$\pm$0.48          & 96.20$\pm$0.51          \\
GraphENS      & 94.10$\pm$1.77          & 94.51$\pm$1.57          & 94.19$\pm$1.78          & 94.88$\pm$1.91          & 94.94$\pm$1.87          & 94.93$\pm$1.94          & 96.26$\pm$1.17          & 96.23$\pm$1.19          & 96.17$\pm$1.15          & 96.37$\pm$1.08          & 96.38$\pm$1.07          & 96.45$\pm$1.09          \\
SNS           & \textbf{95.83$\pm$1.17} & \textbf{95.75$\pm$1.27} & \textbf{95.81$\pm$1.44} & \textbf{95.91$\pm$1.20} & \textbf{95.84$\pm$1.19} & \textbf{95.92$\pm$1.24} & \textbf{96.85$\pm$0.79} & \textbf{96.91$\pm$0.81} & \textbf{96.86$\pm$0.83} & \textbf{96.92$\pm$0.18} & \textbf{96.90$\pm$0.17} & \textbf{96.90$\pm$0.20} \\ \hline
\end{tabular}}
\end{table*}

\subsection{Performance Comparison}
To answer \textbf{Q1}, we compare SNS with 8 baselines on four datasets. The average results with standard deviation are shown in Table \ref{tab_end2end}. From the table, we observe that SNS consistently outperforms baselines in all datasets on different evaluation metrics, which validates the effectiveness of SNS. 
Compared with the vanilla ie-HGCN model, cost-sensitive learning methods (Reweight, Focal loss, and Balanced softmax) perform different levels of growth on four datasets. Reweight and Balanced softmax achieve limited performance increase on ACM and DBLP.
The limited effectiveness of cost-sensitive learning stems from its disregard for topological structures, which hinders its ability to overcome the imbalance problem in HINs.
For re-balanced training strategies (Self-training and cRT), self-training exhibits a significant improvement in classification performance across various metrics and obtains the highest F1 score among 8 baselines on the MAG and ACM datasets. This is attributed to its ability to not only re-balance the minority classes using pseudo labels but also incorporate supplementary information from unlabeled data. Although cRT decouples the feature representation and re-trains the classifier utilizing the balanced labels, its performance is still limited by the lack of label information.  
Among the data augmentation methods, the SOTA method on homogeneous graphs, i.e., GraphENS, exhibits competitive improvement across all evaluation metrics. However, unlike homogeneous graphs, applying GraphENS to HINs can significantly impair the semantics of target nodes, because the neighbor sampling strategy in GraphENS treats all neighbor types equally. Compared with the data augmentation methods, SNS considers the complex semantics of target nodes and avoids introducing noises by semantic-aware and class information constraints, contributing to better classification performance in imbalanced scenarios.

\subsection{Ablation Study}
\label{abl_stu} 
To answer \textbf{Q2} and verify the effectiveness of our proposed Semantic-aware neighbor selection, we evaluate SNS: (i) without semantic-aware neighbor selection but select the neighbor nodes from the neighbors of minority nodes (called ``w/o NS (m)''); (ii) without semantic-aware neighbor selection but select the neighbor nodes from all nodes of neighbor type (called ``w/o NS (a)'').
Furthermore, \textbf{Q3} is answered by evaluating SNS: (iii) without the prototype-based class constraint (called ``w/o PC''); (iv) without the semantic-aware embedding constraint (called ``w/o SE''). Note that SNS is evaluated on the same base model while disabling one component at a time. Table \ref{tab_abl} displays the results of these settings.
Comparing SNS with SNS w/o NS (m), the most significant decline in performance occurs in the dataset MAG. Without the evaluation of the influence on minority classes, the neighbor node candidates are restricted to the minority neighbors, leading to the deficiency of semantics and increasing the over-fitting risk. When the neighbor node candidates are increased to all nodes of the neighbor type, SNS w/o NS (a) shows a decline in performance on different evaluation metrics, which indicates the destruction of the semantics of minority nodes.
Besides the semantic-aware neighbor selection method, SNS also leverages the semantic and class information to guide the embedding generation. 
When ignoring the prototype-based class constraint, SNS w/o PC demonstrates performance drops on all four datasets. As for the semantic-aware embedding constraint, SNS obtains more remarkable gains in all three evaluation metrics when deploying the constraints. Compared to SNS w/o SE, the F1 score of SNS declines by 2.12\% on ACM.  
These obtained results emphasize the significance of constraints, particularly the incorporation of semantics in guiding the representation of synthetic nodes. The semantic constraint facilitates more effective representation by the HGNN and leads to a substantial enhancement in classification performance.

\begin{table*}[t]
\centering
\setlength{\abovecaptionskip}{0.2cm}
\caption{{Experiment results with HGT as the base model. The best performance is bold.}}
 \label{tab_hgt}
 \resizebox{\linewidth}{!}{
\begin{tabular}{l|lll|lll|lll|lll}
\hline
              & \multicolumn{3}{c|}{MAG}                                        & \multicolumn{3}{c|}{ACM}                                        & \multicolumn{3}{c|}{DBLP}                                      & \multicolumn{3}{c}{IMDB}                                              \\ \hline
Method        & F1 score            & ACC                 & BACC                & F1 score            & ACC                 & BACC                & F1 score           & ACC                 & BACC                & F1 score            & ACC                 & \multicolumn{1}{l}{BACC} \\ \hline
Vanilla       & 83.13$\pm$1.54          & 84.12$\pm$1.26          & 83.3$\pm$1.38           & 59.75$\pm$0.53          & 76.77$\pm$3.08          & 62.33$\pm$0.99          & 87.74$\pm$0.79         & 88.49$\pm$0.77          & 88.22$\pm$0.92          & 26.49$\pm$0.92          & 33.31$\pm$0.82          & 40.44$\pm$0.58                \\
Reweight      & 87.45$\pm$2.65          & 87.61$\pm$2.42          & 87.21$\pm$2.64          & 61.45$\pm$1.23          & 80.13$\pm$1.76          & 63.56$\pm$1.29          & 88.46$\pm$1.64         & 88.86$\pm$1.51          & 88.76$\pm$1.44          & 28.77$\pm$1.06          & 34.39$\pm$1.51          & 41.37$\pm$0.80                \\
Focal loss    & 86.91$\pm$2.61          & 87.11$\pm$2.59          & 86.67$\pm$2.81          & 60.51$\pm$1.74          & 79.42$\pm$2.88          & 62.44$\pm$1.14          & 88.11$\pm$1.13         & 88.79$\pm$1.10           & 88.62$\pm$1.10           & 30.37$\pm$1.05          & 35.16$\pm$1.13          & 42.15$\pm$1.39                \\
BS            & 89.88$\pm$1.16          & 90.15$\pm$1.05          & 89.82$\pm$1.06          & 61.92$\pm$1.89          & 81.68$\pm$2.37          & 63.97$\pm$1.24          & 88.97$\pm$0.86         & 89.61$\pm$0.90           & 89.61$\pm$0.76          & 32.61$\pm$1.83          & 37.96$\pm$1.47          & 42.30$\pm$1.55                \\
Self-training & 90.63$\pm$1.18          & 91.09$\pm$1.17          & 90.64$\pm$1.14          & 60.30$\pm$0.67          & 78.94$\pm$0.23          & 62.91$\pm$0.49          & 87.86$\pm$1.15         & 88.51$\pm$1.12          & 88.41$\pm$1.10           & 33.87$\pm$1.85          & 37.05$\pm$1.41          & 41.95$\pm$1.07                \\
cRT           & 86.25$\pm$1.63          & 86.99$\pm$1.27          & 86.39$\pm$1.36          & 62.57$\pm$1.69          & 81.11$\pm$1.46          & 63.82$\pm$1.32          & 87.99$\pm$0.53         & 88.76$\pm$0.51          & 88.44$\pm$0.63          & 27.07$\pm$0.61          & 34.95$\pm$0.59          & 40.98$\pm$0.67                \\
ROS           & 85.50$\pm$3.44           & 85.74$\pm$3.23          & 85.31$\pm$3.49          & 61.13$\pm$0.93          & 81.21$\pm$2.17          & 63.26$\pm$0.74          & 87.94$\pm$0.79         & 88.96$\pm$0.17          & 88.92$\pm$0.62          & 27.33$\pm$1.13          & 34.45$\pm$0.65          & 41.71$\pm$0.56                \\
SMOTE         & 83.71$\pm$2.44          & 84.52$\pm$2.13          & 83.71$\pm$2.30          & 61.02$\pm$1.20          & 81.79$\pm$3.14          & 62.71$\pm$0.92          & 88.71$\pm$0.94         & 88.55$\pm$0.83          & 88.41$\pm$1.00           & 27.73$\pm$1.49          & 33.74$\pm$0.71          & 41.42$\pm$1.44                \\
GraphENS      & 89.23$\pm$1.48          & 89.59$\pm$1.07          & 89.14$\pm$1.01          & 62.45$\pm$1.41          & 81.03$\pm$2.14          & 62.85$\pm$1.54          & 89.86$\pm$0.63         & 90.51$\pm$0.66          & 89.92$\pm$0.71          & 32.42$\pm$1.00           & 37.09$\pm$1.72          & 42.25$\pm$1.23                \\
SNS           & \textbf{91.93$\pm$1.24} & \textbf{91.79$\pm$1.22} & \textbf{91.88$\pm$1.12} & \textbf{63.32$\pm$1.16} & \textbf{82.05$\pm$1.22} & \textbf{64.88$\pm$0.73} & \textbf{90.80$\pm$1.15} & \textbf{91.15$\pm$1.13} & \textbf{91.01$\pm$1.22} & \textbf{34.79$\pm$1.71} & \textbf{38.81$\pm$1.80} & \textbf{42.99$\pm$1.05}       \\ \hline
\end{tabular}}
\end{table*}

\subsection{Varying Imbalance Ratio}

To answer \textbf{Q4}, we test the performance of different baselines with respect to the imbalance ratio on the MAG dataset. The $\textit{imbalance ratio}$ varies as \{0.1, 0.2, 0.3, 0.4, 0.5\}. The experimental results are shown in Table \ref{tab_end2end} and \ref{tab_ibr}. 
As the imbalance ratio increases, all tested methods show good generalization and mitigate the imbalance problem in each imbalance scenario.
Among the cost-sensitive learning methods, Focal loss has the best performance when the imbalance ratio increases. Self-training and GraphENS show stable performance against different imbalance ratios and achieve the highest classification performance among re-balanced training strategies and data augmentation methods, respectively.
Our proposed method SNS could successfully generalize to different imbalance ratios and consistently outperforms other compared methods on all three evaluation metrics. Moreover, the more severe the imbalance degree, the greater the improvement of SNS. 
For instance, as the imbalance ratio decreases, the performance improvement of SNS becomes increasingly significant compared to the base model, with a margin of 0.81\%, 1.31\%, 1.86\%, 4.68\%, and 12.70\% in terms of the F1 score.
The highest improvement under the extreme imbalance ratio scenario further verifies the superiority of SNS, because data augmentation methods have to generate more synthetic nodes and edges which can introduce noise and impair the classification performance. Due to the benefits of semantic preservation and two constraints for synthetic nodes, SNS successfully controls the noise and achieves superior performance.

\subsection{Generalization}
To answer \textbf{Q5}, we test the generalization ability of SNS by deploying it on HGT \cite{hu2020heterogeneous} and compare the 8 baselines. The experimental results are shown in Table \ref{tab_hgt}. 
Similar to ie-HGCN, all compared baselines can alleviate the imbalance issue and improve the classification performance of HGT.
Among the cost-sensitive learning methods, balanced softmax achieves the best performance. Among the re-balanced training strategies, self-training achieves the highest classification performance, while GraphENS achieves the best results among the data augmentation methods.
Despite not considering the semantics among different types of nodes, self-training and GraphENS leverage the topology information of HIN and achieve significant performance gains compared to the other baselines. Compared with all these methods, SNS consistently reaches the highest classification results on all three evaluation metrics.

\begin{figure}[t]
\setlength{\abovecaptionskip}{0.2cm}
\centering
\includegraphics[width=1\linewidth]{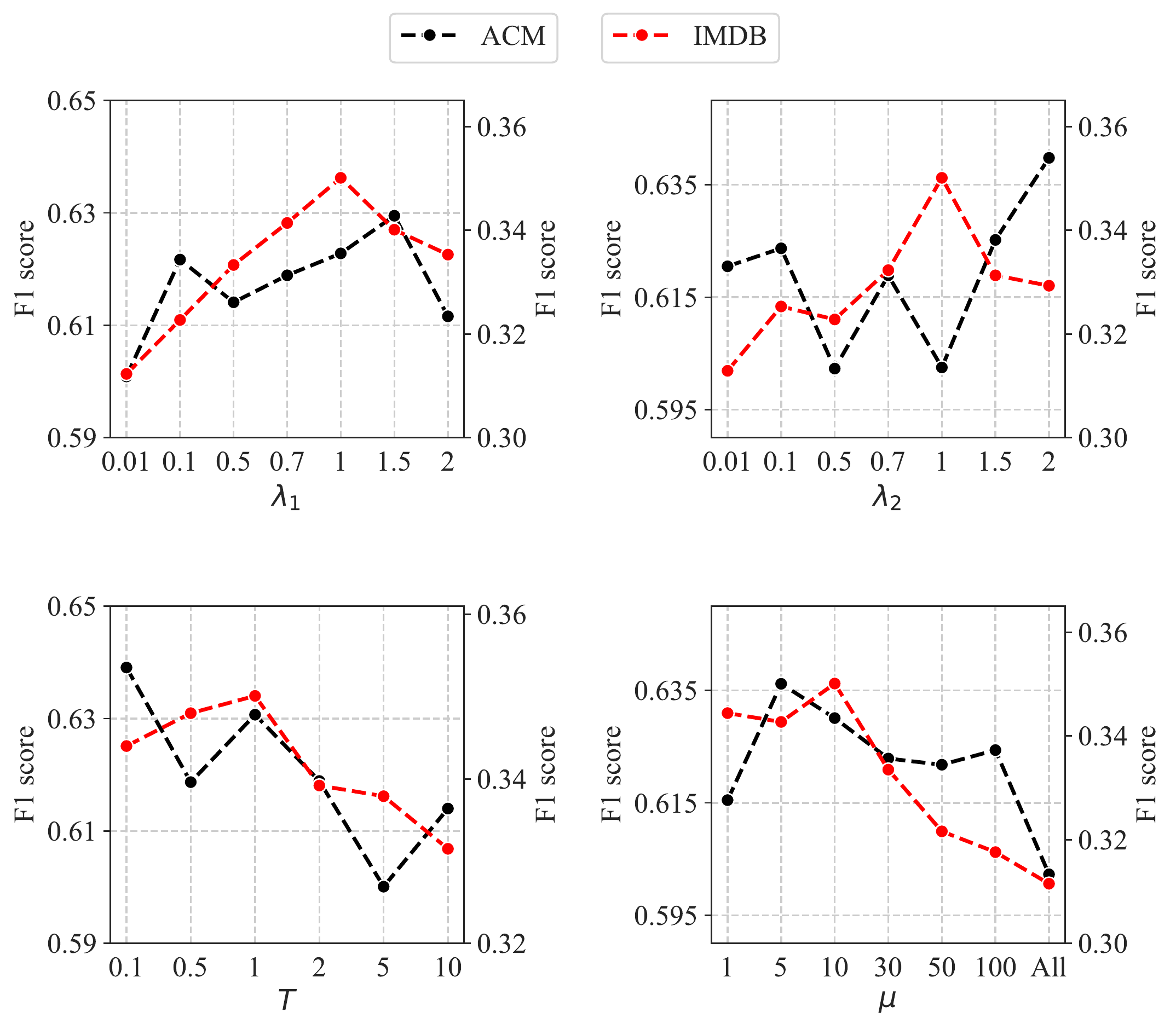}
\caption{Parameter sensitivity results on ACM and IMDB dataset.}
\label{fig_para}
\end{figure}

\subsection{Parameter Sensitivity Analysis}
To analyze the influence of these hyper-parameters and answer \textbf{Q6}, we conduct the parameter sensitivity experiments and show the F1 score performances in terms of hyper-parameters in Fig. \ref{fig_para}. The performances of ACM and IMDB datasets are demonstrated and other datasets are shown similar results.

Firstly, $\lambda_1$ and $\lambda_2$ are quite important which could significantly affect the classification result. $\lambda_1$ should be controlled between 1 and 2 to get better performance. This indicates that the semantic-aware embedding constraint provides an important supervision signal, and increasing the weight of $\mathcal{L}_{sem}$ contributes to a higher quality representation of the synthetic node. A similar trend can be observed in the results in terms of $\lambda_2$. The classification performance decreases first and then increases, and a large weight should be assigned for $\mathcal{L}_{pro}$. 
Following the increase of $T$, the F1 score results show general downward trends. Thus, limiting $T$ to a small value works best.   
Finally, the results in terms of $\mu$ show that increasing the number of neighbor candidates could help enhance the classification performance. But it also introduces more unreliable neighbor nodes for synthetic nodes and disturbs the minority semantics. Especially when introducing all neighbor type nodes, the classification result drops rapidly.

\section{Conclusion}
We present Semantic-aware Node Synthesis (SNS), a data augmentation method for semantic imbalance issue in imbalanced heterogeneous information networks (HINs). SNS can successfully capture the minority semantics within HINs and generate the synthetic nodes to re-balance the class distributions. By employing the semantic-aware embedding constraint and the prototype-based class constraint, SNS effectively utilizes neighbor nodes and class information to guide the HGNN in generating reliable embeddings for synthetic nodes. Extensive experiments demonstrate that SNS achieves high classification performance and well generalization ability for various imbalance scenarios.
\section*{Acknowledgment}
This work is supported by Australian Research Council Future Fellowship (Grant No.FT210100624), Discovery Project (Grant No. DP190101985), and Discovery Early Career Research Award (Grant No.DE200101465 and No.DE230101033).
\bibliographystyle{ACM-Reference-Format}
\bibliography{ref}

\end{document}